\documentclass[journal,twoside,web]{ieeecolor}

\usepackage{etoolbox}
\makeatletter
\@ifundefined{color@begingroup}%
{\let\color@begingroup\relax
\let\color@endgroup\relax}{}%
\def\fix@ieeecolor@hbox#1{%
\hbox{\color@begingroup#1\color@endgroup}}
\patchcmd\@makecaption{\hbox}{\fix@ieeecolor@hbox}{}{\FAILED}
\patchcmd\@makecaption{\hbox}{\fix@ieeecolor@hbox}{}{\FAILED}

\usepackage{tmi}
\usepackage{cite}
\usepackage{amsmath,amssymb,amsfonts}
\usepackage{algorithmic}
\usepackage{graphicx}
\usepackage{textcomp}
\usepackage{multirow} 
\usepackage{booktabs} 
\usepackage{hyperref} 
\usepackage[cmyk]{xcolor}

\def\BibTeX{{\rm B\kern-.05em{\sc i\kern-.025em b}\kern-.08em
    T\kern-.1667em\lower.7ex\hbox{E}\kern-.125emX}}
\begin{document}
\title{
Following the Diagnostic Trace: Visual Cognition-guided Cooperative Network 

for Chest X-Ray Diagnosis
}
\author{Shaoxuan Wu, Jingkun Chen, Chong Ma, Cong Shen, Xiao Zhang, Jun Feng 
\thanks{
This work was supported in part by the National Natural Science Foundation of China under Grant 62403380 and the Shaanxi Province Postdoctoral Science Foundation under Grant 2024BSHSDZZ042.
(Co-first authors: Shaoxuan Wu, Jingkun Chen; Corresponding authors: Xiao Zhang, Jun Feng.)
}
\thanks{
Shaoxuan Wu, Xiao Zhang, and Jun Feng are with the College of Computer Science, Northwest University, Xi’an, China, 710127 (e-mail: wushaoxuan@stumail.nwu.edu.cn; {xiaozhang, fengjun}@nwu.edu.cn). 
}
\thanks{
Jingkun Chen is with the Department of Engineering Science, University of Oxford, Oxford, OX3 7DQ, United Kingdom (e-mail: jingkun.chen@eng.ox.ac.uk).
}
\thanks{
Chong Ma is with the School of Computing and Artificial Intelligence, Southwest Jiaotong University, Chengdu 611756, China (e-mail: machong@swjtu.edu.cn).
}
\thanks{
Cong Shen is with the Department of PET/CT, The First Affiliated Hospital of Xi'an Jiaotong University, Xi'an, 710061, China. (shencong100217@fh.xjtu.edu.cn).
}
}

\maketitle

\begin{abstract}
Computer-aided diagnosis (CAD) has significantly advanced automated chest X-ray diagnosis but remains isolated from clinical workflows and lacks reliable decision support and interpretability. 
Human-AI collaboration seeks to enhance the reliability of diagnostic models by integrating the behaviors of controllable radiologists. 
However, the absence of interactive tools seamlessly embedded within diagnostic routines impedes collaboration, while the semantic gap between radiologists’ decision-making patterns and model representations further limits clinical adoption.
To overcome these limitations, we propose a visual cognition-guided collaborative network (VCC-Net) to achieve the cooperative diagnostic paradigm. VCC-Net centers on visual cognition (VC) and employs clinically compatible interfaces, such as eye-tracking or the mouse, to capture radiologists’ visual search traces and attention patterns during diagnosis.
VCC-Net employs VC as a spatial cognition guide, learning hierarchical visual search strategies to localize diagnostically key regions. A cognition–graph co-editing module subsequently integrates radiologist VC with model inference to construct a disease-aware graph. The module captures dependencies among anatomical regions and aligns model representations with VC-driven features, mitigating radiologist bias and facilitating complementary, transparent decision-making.
Experiments on the public datasets SIIM-ACR, EGD-CXR, and self-constructed TB-Mouse dataset achieved classification accuracies of 88.40\%, 85.05\%, and 92.41\%, respectively. 
The attention maps produced by VCC-Net exhibit strong concordance with radiologists’ gaze distributions, demonstrating a mutual reinforcement of radiologist and model inference.
The code is available at \href{https://github.com/IPMI-NWU/VCC-Net}{https://github.com/IPMI-NWU/VCC-Net}.

\end{abstract}

\begin{IEEEkeywords}
Visual Cognition, Computer-aided Diagnosis, Eye Tracking, Mouse Trajectory
\vspace{-1mm}
\end{IEEEkeywords}

\section{Introduction}
\label{sec:introduction}

\vspace{-2mm}
\begin{figure}[h]
\centerline{\includegraphics[width=8.7cm]{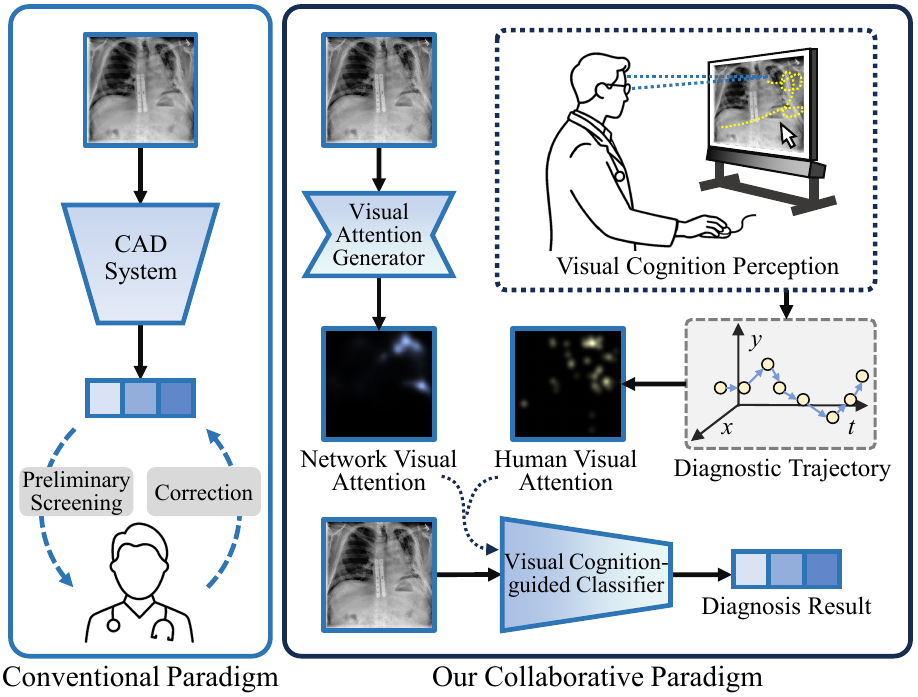}}
\vspace{-1mm}
\caption{The collaborative paradigm (right) leverages radiologists’ visual cognition to bridge radiologists' cognition and model inference. Compared with conventional CAD (left), it enhances the consistency and reliability in clinical decisions.}
\label{fig:Intro}
\vspace{-1.5mm}
\end{figure}

\IEEEPARstart{L}{ung} diseases such as pneumothorax, pneumonia, and tuberculosis impose a substantial global health burden, with tuberculosis alone accounting for more than ten million newly reported cases each year
\cite{kalyanpur2025applications}. Chest X-ray remains a clinically diagnostic modality that radiologists rely on to guide therapeutic decisions. However, the growing volume of imaging examinations places considerable pressure on radiology services, potentially reducing diagnostic accuracy and slowing clinical decision-making.

The rapid advancement of computer-aided diagnosis (CAD) has substantially improved automated chest X-ray diagnosis, offering radiologists effective support for rapid screening \cite{Yifan2020MIA,zhang2024anatomy}. Despite achievements, most models still rely on end-to-end, data-driven paradigms and function as isolated computational models with limited integration into clinical workflows \cite{zhao2022diagnose}. 
In addition, the models are susceptible to non-clinical factors \cite{Shortcut} and demonstrate limited interpretability, factors that weaken radiologists’ trust \cite{EGViT}. These limitations hinder models from becoming trustworthy decision-making partners and restrict their widespread adoption in automated chest X-ray diagnosis.

In recent years, the reassessment of the relationship between computer-aided diagnosis (CAD) and radiologists has emerged as a key research direction in chest X-ray diagnosis. Rather than functioning solely as a decision-support role, the new generation of models promotes collaboration, enabling radiologists and models to participate in the diagnostic process jointly \cite{Pershin2022JBHI}. By introducing radiologist decision-making patterns, human-AI collaboration enhances the transparency and reliability of diagnostic results while maintaining automation efficiency \cite{wang2024interactive}. Nevertheless, the lack of interactive tools that integrate seamlessly into routine clinical workflows constrains effective collaboration, and the semantic discrepancy between radiologists’ decision-making patterns and model representations continues to impede broader clinical adoption.

Radiologists’ visual cognition (VC) during image reading represents a natural bridge between radiologist reasoning and model inference. VC captures the decision-making process of radiologists and can be captured through a clinically compatible manner—such as eye-tracking or mouse—without interrupting standard diagnostic routines  \cite{GazeMTL,GANet}. The cognitive process typically follows a hierarchical search strategy, beginning with global structural scanning and progressing toward localized inspection of suspicious lesions \cite{RadioTransformer}.
Fine-grained abnormalities such as nodules often elude models, while radiologists’ attention patterns compensate for these gaps. Whereas model-generated cues reduce biases caused by fatigue or subjectivity. The bidirectional synergy between the two establishes a solid foundation for developing an effective collaborative diagnostic paradigm.

Building upon these insights, a visual cognition-guided cooperative network (VCC-Net) is proposed to implement an efficient collaborative diagnostic paradigm. VCC-Net comprises two principal modules: The visual attention generator (VAG) learns radiologists' hierarchical visual search strategies, progressively refining focus from global to local regions to dynamically localize clinically relevant areas. The visual cognition-guided classifier (VCC) employs a cognition–graph co-editing module to integrate radiologists’ VC patterns with attention maps produced by the model. The resulting graph structure encodes pathological semantics and captures interdependencies among anatomical regions, enabling more comprehensive diagnostic reasoning. As illustrated in Fig.\ref{fig:Intro}, radiologists’ fixation provides spatial constraints that guide the model to concentrate on clinically pertinent regions, while the model’s attention maps assist radiologists in detecting subtle findings or correcting subjective deviations. Such mutual reinforcement facilitates complementary decision-making and cognitive alignment throughout the diagnostic process. Experimental evaluations confirm that VCC-Net delivers consistent improvements in both diagnostic accuracy and reliability.

The main contributions of this paper are as follows:
\begin{itemize}
\item We propose VCC-Net, which learns and integrates radiologists’ visual search traces and attention patterns through the VAG and VCC modules, respectively, to achieve a collaborative and complementary diagnostic paradigm.
\item The VAG designed to emulate the hierarchical search strategy characteristic of radiologists, combines the global contextual modeling capabilities with the localized feature extraction strengths. By leveraging VC, it captures radiologists' visual search behavior and generates attention maps that emphasize clinically significant regions.
\item The VCC employs a cognition–graph co-editing module to integrate VC for constructing a disease-aware graph. The module captures inter-regional dependencies among anatomical regions and aligns model representations with VC features, mitigating radiologist bias and facilitating transparent diagnostic decisions.
\item Extensive experiments on two public gaze datasets, SIIM-ACR and EGD-CXR, along with the self-constructed TB-Mouse dataset, demonstrate that the VCC-Net outperforms current state-of-the-art methods in both diagnostic accuracy and clinical interpretability.
\end{itemize}

\vspace{-3mm}
\section{Related Work}

\subsection{Learning Radiologists' Attention for Medical Imaging Analysis}
In recent years, attention mechanisms have gained widespread adoption in medical image analysis. Inspired by radiologists' VC, attention mechanisms guide models to prioritize key regions in images \cite{TPAMI_Bai}.
Pioneering studies augmented convolutional neural
network (CNN) architectures with attention modules, sharpening their focus on diagnostically pertinent regions \cite{MIA_Guo,TMI_Gu}.
These methods have integrated spatial and channel attention-through techniques, such as adaptive aggregation and collective spatial attention more precisely highlight abnormal regions.

Self-attention is inspired by the cognitive processes of the brain and aims to emulate the ability to selectively concentrate on salient information through dynamic weight allocation. It has gained significant attention due to its effectiveness in handling long-range dependencies.
Recent methods have improved contextual representation by introducing global-local feature interaction mechanisms \cite{TMI_He} or by integrating the local feature extraction capabilities of CNN with the long-range dependency modeling strengths of Transformers \cite{PR_Xiayu}. In addition, deformable Transformers have been employed to further refine attention to clinically significant regions within medical images \cite{MICCAI_Xie}.

The key advantage of attention mechanisms is their ability to simulate attention by selectively focusing on critical information, thus enhancing feature recognition. However, it remains uncertain whether the attention learned by models accurately reflects attention. VC, which encompasses attentional behavior during image interpretation, offers a promising avenue for guiding feature selection and improving model interpretability—an essential factor in fostering clinical trust in automated diagnostic models \cite{EGViT,TPAMI_Bai}.
VC is expected to mitigate the mismatch between model focus and the clinical judgment process by aligning attention mechanisms with radiologists-driven visual cues.

\begin{figure*}[ht]
\centerline{\includegraphics[width=17cm]{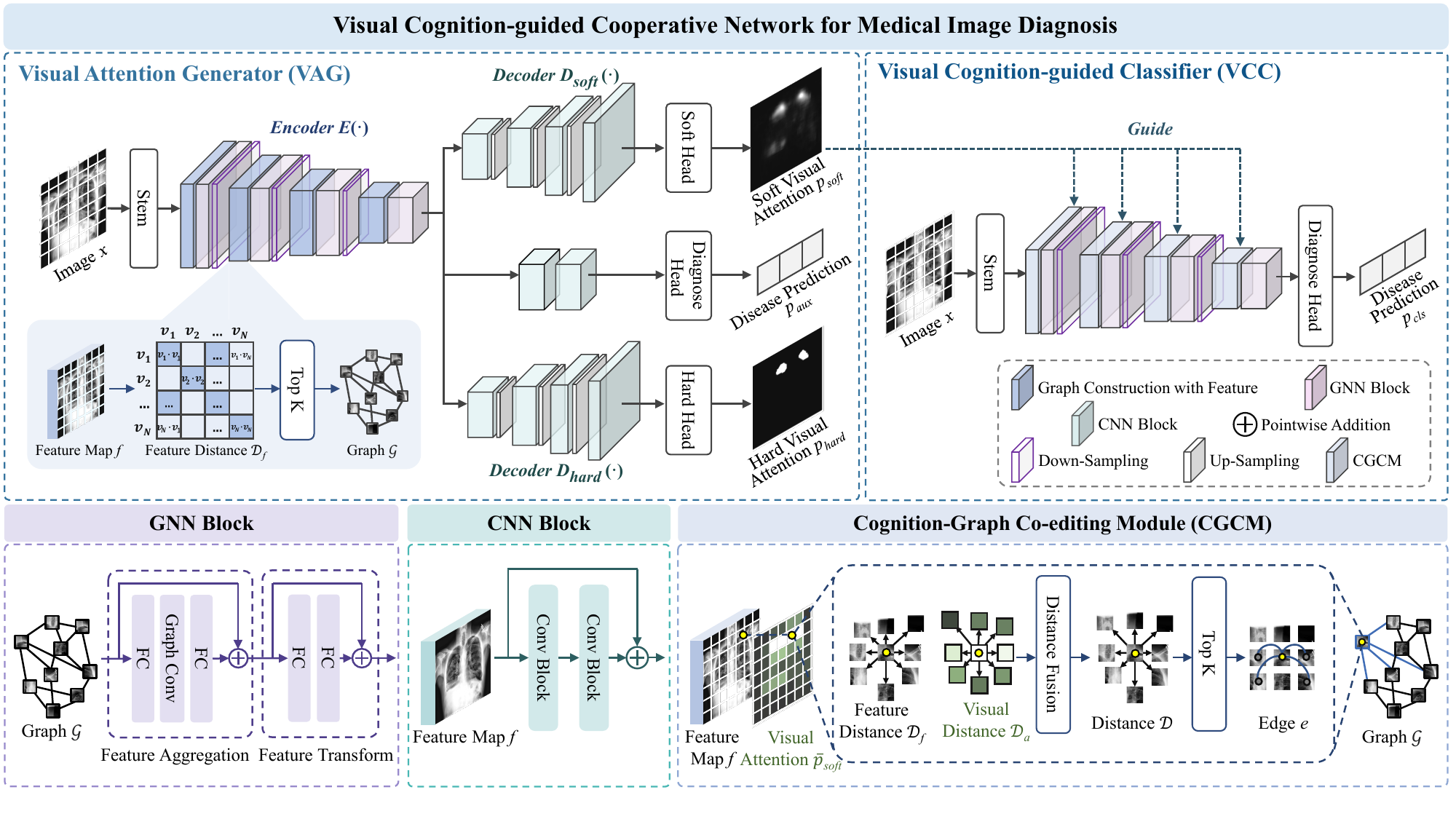}}
\vspace{-1.5mm}
\caption{The proposed VCC-Net comprises two main components: (1) The VAG employs GNN and CNN to model both global and local visual search patterns of radiologists, and supervises learning through three pathways. The VAG generates radiologist-like visual attention based on the input medical images. (2) The VCC leverages radiologists' VC to construct a graph structure and align visual and feature differences across regions, ultimately producing diagnostic outputs.}
\vspace{-3mm}
\label{fig:Method_VCNet}
\end{figure*}

\vspace{-2mm}
\subsection{Integrating Visual Cognition into Collaborative Medical Image Analysis}
Radiologists' VC is reflected in the distribution of their attention across medical images and radiologists' decision-making processes \cite{GazeMTL,RadioTransformer,Wang2025TMI}.
VC can be effectively recorded by tracking radiologists' gaze or mouse trajectories. 
Gaze record visual fixation points, highlighting regions most relevant to diagnosis \cite{MIA_Alsharid}, while mouse trajectories track user activity and stay points at varying granularities during interaction \cite{CHB_Katerina,AQD_Horwitz}.
This data collection framework preserves the integrity of clinical workflows while providing comprehensive, multidimensional data support for CAD.

Recent studies have integrated radiologists’ VC into CAD. 
For example, GA-Net was proposed by Wang et al. \cite{GANet}, where gaze maps and attention consistency were used to guide the network’s focus, thereby improving diagnostic accuracy from knee X-rays. 
EG-ViT \cite{EGViT} uses gaze information to mask irrelevant background features, reducing shortcut learning. 
GazeGNN \cite{GazeGNN} embeds gaze into the graph neural network (GNN) input to enhance inference robustness, also improving interpretability by constructing graph nodes from image patches.
Xie et al. \cite{TMI_Xie} introduced a medical image segmentation framework that integrates eye-tracking information as weak supervision, improving performance under limited data conditions.
Additionally, Gajos et al. \cite{MD_Gajos} developed Hevelius, a rapid mouse trajectory test, to support objective assessments of movement disorders in clinical trials.
Li et al. \cite{CVPR_Li} proposed AG-CNN, an attention-based CNN that addresses redundancy in retinal fundus images, improving glaucoma detection.
Additionally, Hevelius was developed by Gajos et al. \cite{MD_Gajos} as a rapid mouse trajectory test, designed to facilitate objective movement disorder assessments in clinical trials. 
AG-CNN was proposed by Li et al. \cite{CVPR_Li}, an attention-based CNN that addresses redundancy in retinal fundus images, thereby enhancing glaucoma detection.

Existing studies have validated the potential of VC to enhance diagnostic performance in CAD \cite{wu2024MICCAI}. Current mainstream methods primarily guide models to focus on specific regions by embedding or masking background information, but have yet to explore the visual search patterns and anatomical associations inherent in VC.
Graph-based methods offer a promising strategy for modeling spatial and semantic relationships in medical images. By leveraging VC-informed relational modeling, it is possible to capture inter-regional correlations.
Aligning graph structures with cognitive strategies enables the learning of clinical decision-making processes, thereby contributing to improved performance and interpretability.

\vspace{-2mm}
\section{Method}
The VCC-Net aims to improve chest X-ray diagnosis performance and transparency by leveraging radiologists' visual cognition (VC), \textit{i.e.}, gaze or mouse trajectory, during film reading.
As shown in Fig.\ref{fig:Method_VCNet}, the proposed VCC-Net consists of the VAG and the VCC. 
The VAG learns visual patterns from radiologists and generates corresponding visual attention from medical images. 
The VCC leverages visual attention to construct graph structures and align feature differences and visual differences, guiding the model to focus on disease-related regions, mirroring radiologists' diagnostic processes. 
The following sections describe the data collection and label generation in Section \ref{section:Method_A}, the network architecture of the VAG in Section \ref{section:Method_B}, and the architecture of the VCC in Section \ref{section:Method_C}.

\begin{figure*}[ht]
\centerline{\includegraphics[width=17cm]{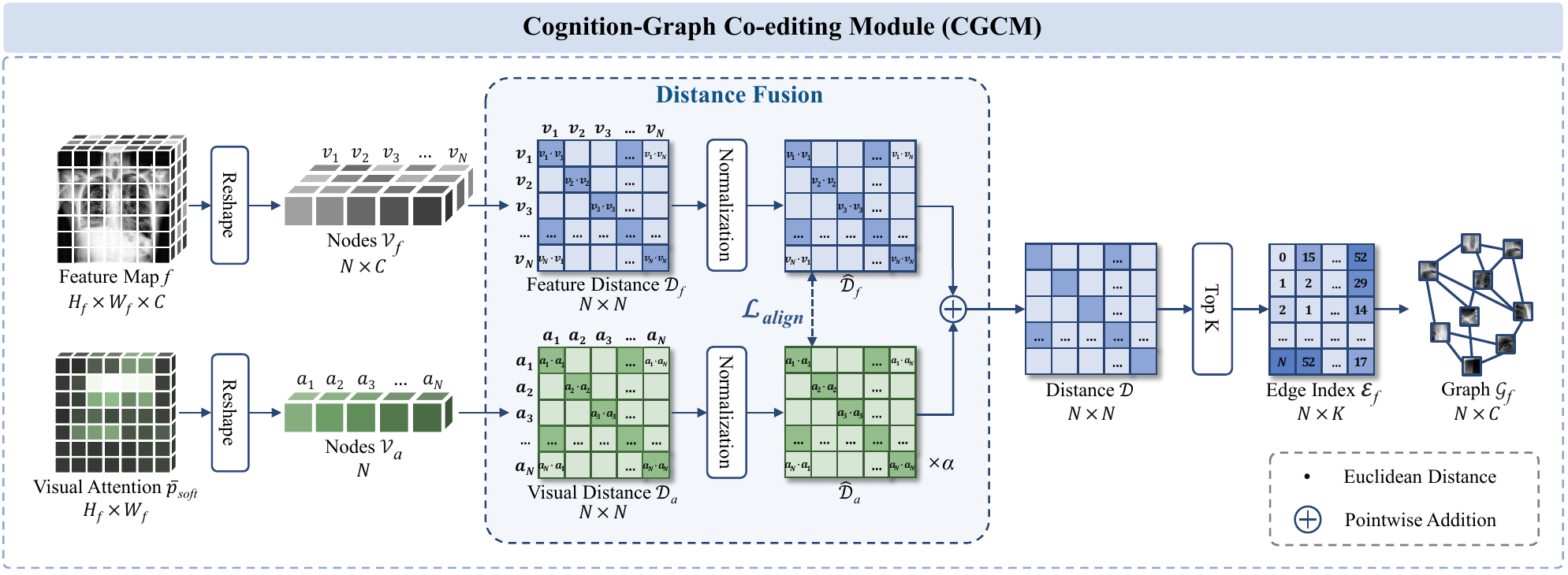}}
\vspace{-1.5mm}
\caption{The CGCM aligns feature and visual distances for each image node by incorporating radiologists' VC, optimizing the feature space structure. With the addition of soft visual attention. The CGCM also guides the network in constructing a graph structure concentrated on high-attention regions, establishing a robust model aligned with radiologists' cognition.}
\vspace{-3mm}
\label{fig:Method_VCGC.pdf}
\end{figure*}

\vspace{-2mm}
\subsection{Data Collection and Label Generation}
\label{section:Method_A}
The eye tracker automatically collects the gaze data and can be seamlessly integrated into the radiologist's workflow. 
For the gaze points in the gaze datasets SIIM-ACR \cite{SIIMACR} and EGD-CXR \cite{EGDCXR,MIMIC}, we process gaze maps based on the data preprocessing steps from \cite{EGViT}, which are referred to as soft visual attention $y_{soft}$ in this paper. 
These reflect the areas of attention of the radiologists.

Compared to gaze data, mouse trajectory data are easier to collect and can be obtained using standard input devices, such as a mouse. 
To this end, we developed a system that automatically captures mouse trajectories during reading. 
As the radiologist interacts with the images, the system tracks the mouse position, recording a set of points $M_{points} = \{p_1, p_2, \cdots, p_m\}$, where each point $p$ represents the mouse position at a particular moment.

After data collection, two key post-processing steps are performed. First, similar to the fixation extraction process used with gaze, we apply the I2MC \cite{I2MC} algorithm to identify stay points $M_{stay}=\{p_1, p_2, \cdots, p_n\}$, which represent regions of prolonged attention during reading.
Next, a 2D Gaussian kernel with a $150$-pixel radius and a $25$-pixel sigma is applied to the stay points, generating a soft visual attention $y_{soft}\in\mathbb{R}^{H\times W}$, where $H$ and $W$ are the image height and width, respectively. 
The visual attention $y_{soft}$ reflects the radiologist's focus, with higher values indicating areas that attracted more attention. 
The same method is applied to gaze datasets to generate corresponding attention.

\vspace{-2mm}
\subsection{Visual Attention Generator}
\label{section:Method_B}
The VAG aims to learn radiologists' VC from visual attention generated through gaze or mouse trajectory. 
By combining the global modeling capability of GNNs and the local feature extraction capability of CNNs, the VAG generates visual attention corresponding to an image. This structure mimics the radiologist's diagnostic process, where they first identify suspicious regions globally and then focus on local regions to identify anomalies like small nodules or exudation.

VAG consists of an encoder $E(\cdot)$ and two decoders, $D_{soft}(\cdot)$ and $D_{hard}(\cdot)$. 
The encoder includes a feature-based graph construction layer and a GNN block for extracting global features. 
Specifically, the input image $x\in\mathbb{R}^{H\times W\times 1}$  is first downsampled through a stem layer to $\frac{H}{4}\times \frac{W}{4}\times C$ , where $C$ represents the feature dimension. 
The graph construction layer transforms features at each position into a node $v\in\mathbb{R}^{C}$, forming a node-set 
$\mathcal{V}=\{v_1,v_2,\cdots,v_N\}$, 
where $N=\frac{H}{4} \ast  \frac{W}{4}$. 
Based on the $\mathcal{V}$, the distance matrix $\mathcal{D}_f\in\mathbb{R}^{N\times N}$ is calculated as follows:
\begin{equation}
    \mathcal{D}_f^{i,j} = \parallel v^i - v^j \parallel_2,
    \label{eq_1}
\end{equation}
where $i$ and $j$ denote the index of nodes.
Using the distance matrix $\mathcal{D}_f$, the nearest neighbors of each node are identified, forming a graph structure $\mathcal{G}=\{\mathcal{V}, \mathcal{E}\}$, where $\mathcal{E}$ denotes the edges between nodes. 
The graph can be represented by the feature vector $X\in\mathbb{R}^{N\times C}$.
The GNN block performs feature aggregation and transformation operations, which can be formulated as:
\begin{equation}
    X_1 = FC_2(GC(FC_1(X))) + X,
    \label{eq_2}
\end{equation}
\begin{equation}
   Y = FC_4(FC_3(X_1)) + X_1,
    \label{eq_3}
\end{equation}
where $FC(\cdot)$ denotes fully connected layers and $GC(\cdot)$ indicates max-relative graph convolution \cite{MRGC}.

Both decoders consist of four CNN blocks for local feature extraction. The CNN block is defined as:
\begin{equation}
   Z = Conv_2(Conv_1(Y)) + Y,
    \label{eq_4}
\end{equation}
where $Conv(\cdot)$ include $3\times3$ convolution layers, batch normalization, and ReLU activation. 
Skip connections between encoder and decoder blocks at the same scale allow integration of global and local information.
After through $D_{soft}(\cdot)$ and the soft head, the soft visual attention $p_{soft}$ is generated.
Similarly, the hard visual attention $p_{hard}$, consisting of binary values indicating regions of high attention, is generated after passing through $D_{hard}(\cdot)$ and the hard head.
Finally, the predicted category $p_{aux}$ is obtained via the diagnose head. Both $p_{aux}$ and $p_{hard}$ as complementary information to improve $p_{soft}$ generation quality. The loss function for VAG is defined as:
\begin{equation}
\begin{aligned}
   \mathcal{L}_{VAG} &= 
   \mathcal{L}_{soft} + \mathcal{L}_{hard} + \mathcal{L}_{aux} \\&
    = \mathcal{L}_{MSE}(p_{soft}, y_{soft}) + \mathcal{L}_{Dice}(p_{hard}, y_{hard}) \\& ~~~ +  \mathcal{L}_{CE}(p_{aux}, y_{cls}),
    \label{eq_5}
\end{aligned}
\end{equation}
where $y_{soft}$ is the ground truth soft visual attention, $y_{cls}$ is the category label, and $y_{hard}$ is a hard label obtained by thresholding $y_{soft}$ $($\textit{i.e.}, $y_{hard} = \mathbb{I}(y_{soft} > threshold)$$)$.
The indicator function $\mathbb{I}(\cdot)$ evaluates to 1 when the probability exceeds the threshold and 0 otherwise.
The $\mathcal{L}_{MSE}$, $\mathcal{L}_{Dice}$, and $\mathcal{L}_{CE}$ correspond to mean squared error loss, Dice loss, and cross-entropy loss, respectively.

\vspace{-2mm}
\subsection{Visual Cognition-guided Classifier}
\label{section:Method_C}
The VCC integrates visual cognition through the cognition-graph co-editing module (CGCM) layer to guide the model's focus on task-relevant areas. 
By leveraging radiologists' attention as supervisory signals, VCC aligns visual and feature differences across regions, enabling the model to learn feature representations consistent with radiologists' cognition. As shown in Fig.\ref{fig:Method_VCNet}, VCC consists of the CGCM and GNN blocks, which share a similar structure to the VAG encoder, but differ in graph construction.

For a feature map $f\in\mathbb{R}^{H_f\times W_f\times C}$, it is treated as a node $\mathcal{V}_f=\{v_1, v_2, \dots, v_N\}$, where $N=H_f\ast W_f$. 
The prediction of soft visual attention map $p_{soft}$ is downsampled to the feature map size, resulting in $\bar{p}_{soft}  \in \mathbb{R}^{H_f \times W_f}$, which is also converted into a point set $\mathcal{V}_a=\{a_1, a_2, \dots, a_N\}$.
Based on the $f$ and $\bar{p}_{soft}$, we compute two distance matrices, $\mathcal{D}_f$ and $\mathcal{D}_a$, which are normalized via min-max scaling to obtain $\hat{\mathcal{D}}_f$ and $\hat{\mathcal{D}}_a$.
The $\hat{\mathcal{D}}_f(i, j)$ and $\hat{\mathcal{D}}_a(i, j)$ represent the differences between the $i$-th and $j$-th nodes in feature and visual attention space, respectively.
By aligning $\hat{\mathcal{D}}_f$ and $\hat{\mathcal{D}}_a$, we help the model learn feature representations that align with the radiologist's attention.
The alignment loss is defined as:
\begin{equation}
   \mathcal{L}_{align} = \mathcal{L}_{MSE}(\hat{\mathcal{D}}_f, \hat{\mathcal{D}}_a).
    \label{eq_6}
\end{equation}

Additionally, the distance matrices $\hat{\mathcal{D}}_f$ and $\hat{\mathcal{D}}_a$ are fused to form the final distance matrix $\mathcal{D} = \hat{\mathcal{D}}_f + \alpha\hat{\mathcal{D}}_a$.
Using $\mathcal{D}$, we identify the $k$-nearest neighbors for each node, forming an edge set $\mathcal{E}_f$ and constructing the graph structure $\mathcal{G}_f=\{\mathcal{V}_f, \mathcal{E}_f\}$.
During the fusion stage, feature and visual distances provide complementary information to build a more robust graph structure.
As shown in Fig.\ref{fig:Result_graph}, distance fusion effectively eliminates connections to areas unrelated to the disease. 
The loss function for VCC is:
\begin{equation}
    \mathcal{L}_{VCC} = \mathcal{L}_{CE}(p_{cls}, y_{cls}) + \lambda_{align}~\mathcal{L}_{align},
    \label{eq_7}
\end{equation}
where $\mathcal{L}_{align}$ is a balancing coefficient for the alignment loss. 
$p_{cls}$ represents the diagnostic prediction results of VCC, which are obtained through several CGCM and GNN blocks, down-sampling, and the diagnose head.
The total loss for VCC-Net is the weighted sum of VAG and VCC losses:
\begin{equation}
   \mathcal{L} = \mathcal{L}_{VCC} + \lambda_{VAG}~  \mathcal{L}_{VAG},
    \label{eq_8}
\end{equation}
where $\lambda_{VAG}$ is the balancing coefficient for the VAG loss.

\begin{table}[!t]  \small
    \centering
    \caption{ 
    Quantitative comparison between our method and other approaches on the SIIM-ACR dataset. Bold values represent the best results. \\
    }
    \vspace{-3mm}
    
    \setlength{\tabcolsep}{8pt}
    \begin{tabular}{p{80pt}|p{25pt}p{25pt}p{25pt}}
    \hline
    
    Method    & \textbf{Acc$\uparrow$}  & \textbf{AUC$\uparrow$}  & \textbf{F1$\uparrow$}  \\
    
    \hline

    \textcolor{blue}{$\spadesuit$} ResNet18 \cite{ResNet}
    & 83.20  & 82.35 & 85.25 \\
    \textcolor{blue}{$\spadesuit$} ResNet50 \cite{ResNet}
    & 84.00  & 85.81 & 83.63 \\
    \textcolor{blue}{$\spadesuit$} ResNet101 \cite{ResNet}
    & 84.40  & 86.13 & 81.08 \\
    \textcolor{blue}{$\spadesuit$} ViT \cite{ViT}
    & 83.60  & 84.16 & 83.77 \\
    \textcolor{blue}{$\spadesuit$} SwinT \cite{SwinT}
    & 84.40  & 83.31 & 83.69 \\
    \textcolor{blue}{$\spadesuit$} ViG \cite{ViG}
    & 83.20  & 84.89 & 82.81 \\

    \hline
   
    \textcolor{green}{$\blacklozenge$} M-SEN \cite{MSEN}
    & 84.80 & 85.93 & 84.03 \\
    \textcolor{green}{$\blacklozenge$} EML-Net \cite{EMLNet}
    & 85.20 & 83.65 & 85.25 \\
    \textcolor{green}{$\blacklozenge$} DeepGaze \cite{DeepGaze}
    & 84.40 & 85.89 & 84.84 \\
    \textcolor{green}{$\blacklozenge$} GA-Net18 \cite{GANet}
    & 84.80 & 71.26 & 83.71 \\
    \textcolor{green}{$\blacklozenge$} GA-Net50 \cite{GANet}
    & 83.20 & 70.25 & 82.35 \\
    \textcolor{green}{$\blacklozenge$} GA-Net101 \cite{GANet}
    & 84.80 & 72.68 & 84.03 \\
    \textcolor{green}{$\blacklozenge$} EG-ViT \cite{EGViT}
    & 85.60 & 75.30 & 85.14 \\
    \textcolor{green}{$\blacklozenge$}  \textbf{Ours}  
    & \textbf{88.40} & \textbf{86.12} & \textbf{88.08}   \\

    \hline
    
    \textcolor{teal}{$\blacksquare$} GazeGNN \cite{GazeGNN}
    & 85.60  & 85.16 & 85.60 \\
   
    \hline
    \end{tabular}
    \label{tab:tab_SIIM}
    \vspace{-3mm}
\end{table}

\begin{table}[!t] \small
    \centering
    \caption{ 
    Quantitative comparison between our method and other approaches on the EGD-CXR dataset. Bold values represent the best results. \\
    }
    \vspace{-3mm}
    \setlength{\tabcolsep}{8pt}
    \begin{tabular}{p{85pt}|p{25pt}p{25pt}p{25pt}}
    \hline
    
    Method    & \textbf{Acc$\uparrow$}  & \textbf{AUC$\uparrow$}  & \textbf{F1$\uparrow$}  \\
    
    \hline

    \textcolor{blue}{$\spadesuit$} ResNet18 \cite{ResNet}
    & 71.96  & 85.02 & 72.17 \\
    \textcolor{blue}{$\spadesuit$} ResNet50 \cite{ResNet}
    & 72.90  & 86.43 & 72.45 \\
    \textcolor{blue}{$\spadesuit$} ResNet101 \cite{ResNet}
    & 74.77  & 85.63 & 74.90 \\
    \textcolor{blue}{$\spadesuit$} ViT \cite{ViT}
    & 70.09  & 85.43 & 69.19 \\
    \textcolor{blue}{$\spadesuit$} SwinT \cite{SwinT}
    & 71.96  & 86.44 & 74.39 \\
    \textcolor{blue}{$\spadesuit$} ViG \cite{ViG}
    & 75.70  & 85.71 & 75.62 \\

    \hline
   
    \textcolor{green}{$\blacklozenge$} M-SEN \cite{MSEN}
    & 78.50 & 84.27 & 77.45 \\
    \textcolor{green}{$\blacklozenge$} EML-Net \cite{EMLNet}
    & 77.57 & 87.33 & 75.47 \\
    \textcolor{green}{$\blacklozenge$} DeepGaze \cite{DeepGaze}
    & 74.77 & 87.87 & 69.40 \\
    \textcolor{green}{$\blacklozenge$} GazeMTL \cite{GazeMTL}
    & 78.50 & 88.70 & 77.90 \\
    \textcolor{green}{$\blacklozenge$} IAA \cite{IAA}
    & 78.50 & 90.00 & 77.60 \\
    \textcolor{green}{$\blacklozenge$} EffNet+GG \cite{EffNet}
    & 77.57 & 88.80 & 77.00 \\
    \textcolor{green}{$\blacklozenge$} GA-Net18 \cite{GANet}
    & 77.57 & 86.13 & 77.47 \\
    \textcolor{green}{$\blacklozenge$} GA-Net50 \cite{GANet}
    & 78.50 & 86.43 & 77.91 \\
    \textcolor{green}{$\blacklozenge$} GA-Net101 \cite{GANet}
    & 79.44 & 86.42 & 79.20 \\
    \textcolor{green}{$\blacklozenge$} EG-ViT \cite{EGViT}
    & 77.57 & 85.53 & 77.42 \\
    \textcolor{green}{$\blacklozenge$}  \textbf{Ours}  
    & \textbf{85.05} & 91.52 & \textbf{84.87}   \\

    \hline
    
    \textcolor{teal}{$\blacksquare$} GazeGNN \cite{GazeGNN}
    & 83.18 & \textbf{92.30} & 82.30 \\
   
    \hline
    \end{tabular}
    \label{tab:tab_EGD}
    \vspace{-4mm}
\end{table}

\vspace{-1.5mm}
\section{Experiments}
\subsection{Dataset and Evaluation Metrics}
To evaluate the performance of VCC-Net, we used three datasets: two public gaze datasets, SIIM-ACR \cite{SIIMACR} and EGD-CXR \cite{EGDCXR,MIMIC}, and the self-constructed TB-Mouse dataset. The SIIM-ACR dataset includes 1,170 chest X-ray images, with corresponding gaze data; 268 of these images are pneumothorax.
The EGD-CXR dataset comprises 1,083 chest X-ray images sourced from the MIMIC-CXR dataset \cite{MIMIC}, categorized into three classes: normal, congestive heart failure, and pneumonia. Each image includes gaze data. 
The TB-Mouse dataset contains 1,000 training images and corresponding mouse trajectories, collected as radiologists read the images, along with 2,200 test images. 
These images are classified into two categories: tuberculosis and normal. 
An experienced radiologist collected the mouse trajectory data using the system described in Section \ref{section:Method_A}.

\begin{figure*}[ht]
\centerline{\includegraphics[width=17cm]{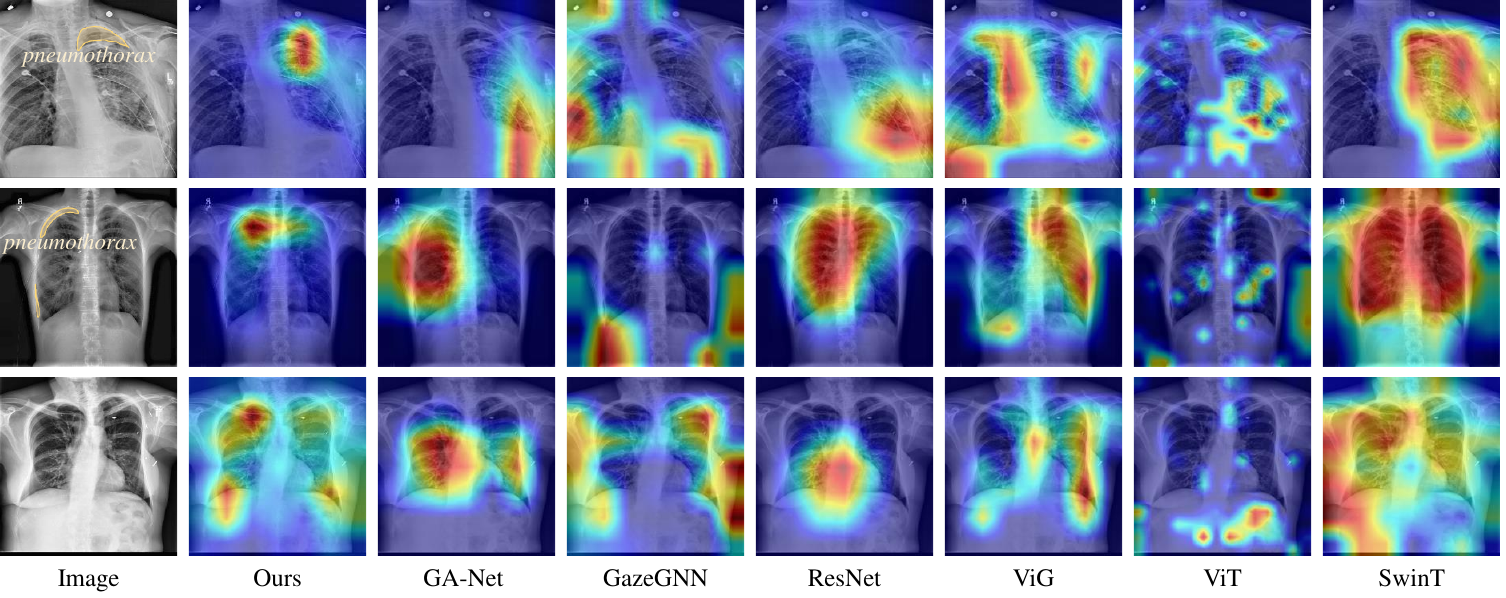}}
\vspace{-1.8mm}
\caption{Visualization of attention maps, generated with Grad-CAM, comparing different methods on the SIIM-ACR dataset. In the first column, the pneumothorax region is highlighted in yellow. The red areas in the attention maps reflect the focus regions of each network. The second column shows that VCC-Net achieves superior anomaly localization by accurately focusing on the pneumothorax region in comparison to other models.}
\label{fig:Result_SIIM_grad}
\vspace{-1mm}
\end{figure*}

To evaluate the performance in diagnosis, we employed accuracy (Acc), area under the receiver operating characteristic curve (AUC), and F1 score (F1) as evaluation metrics.
All experiments were conducted on an NVIDIA 3080 Ti GPU (12GB) using PyTorch. 
We initialized the network with pre-trained model weights from ImageNet \cite{ImageNet} and trained it using the Adam optimizer.
The training process consisted of 200 epochs, with a learning rate of $2 \times 10^{-4}$ and a batch size of 8. 
In Eq.\ref{eq_7} and Eq.\ref{eq_8}, the values of $\lambda_{align} = 0.5$ and $\lambda_{VAG} = 0.5$ were used.
Also in distance fusion, the hyperparameter $\alpha$ is set to 2.0.

\vspace{-3mm}
\subsection{Results on Gaze Dataset}
In our experiments on the gaze datasets, we categorized methods into three groups: 1) Methods without VC \textcolor{blue}{$\spadesuit$}~: ResNet \cite{ResNet}, Vision Transformer \cite{ViT}, Swin Transformer \cite{SwinT}, Vision GNN \cite{ViG}; 2) Methods with VC during training \textcolor{green}{$\blacklozenge$}~: M-SEN \cite{MSEN}, EML-Net \cite{EMLNet}, DeepGaze \cite{DeepGaze}, GA-Net \cite{GANet}, EG-ViT \cite{EGViT}; 3) Methods with VC during inference \textcolor{teal}{$\blacksquare$}~: GazeGNN \cite{GazeGNN}.

\begin{figure*}[ht]
\centerline{\includegraphics[width=17cm]{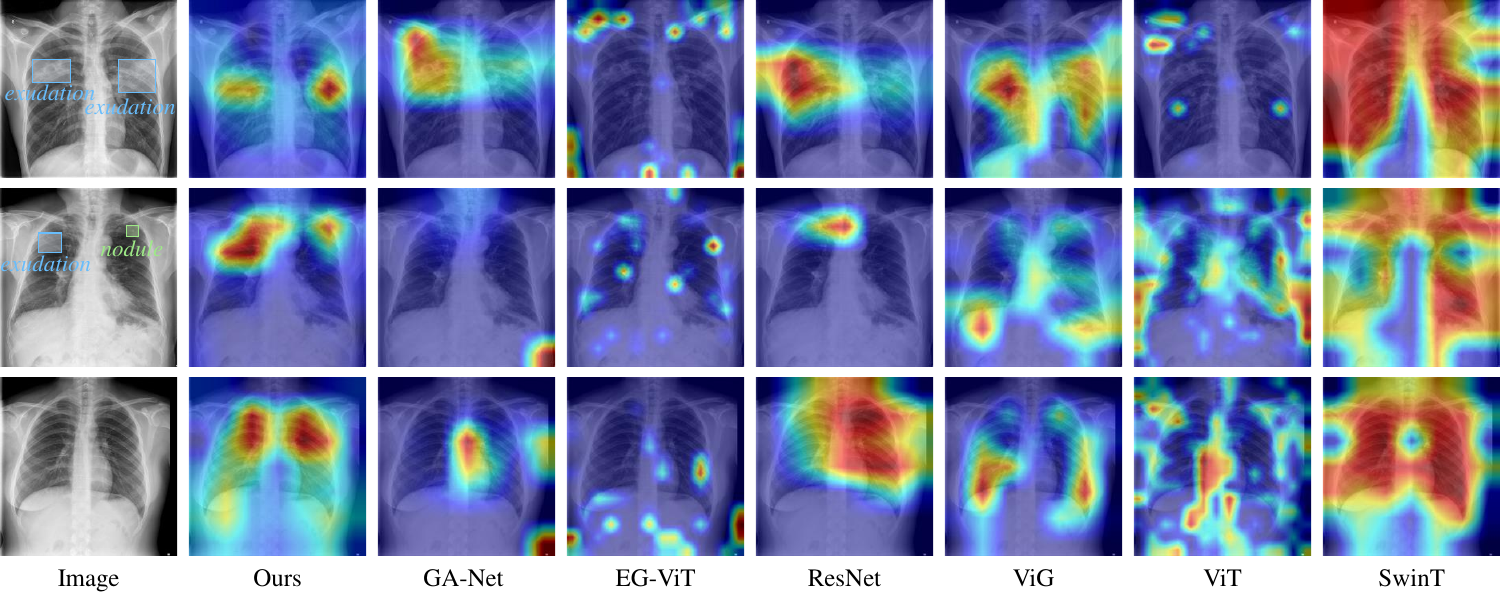}}
\vspace{-1.8mm}
\caption{Visualization of attention maps obtained from Grad-CAM comparing methods on the TB-Mouse dataset. In the first column, regions with exudation and nodules are marked with bounding boxes. Red areas indicate the focused regions in each network's attention map. Compared to other methods, VCC-Net accurately localizes lesion areas.}
\label{fig:Result_TB_grad}
\vspace{-3.5mm}
\end{figure*}

\begin{figure*}[ht]
\centerline{\includegraphics[width=17cm]{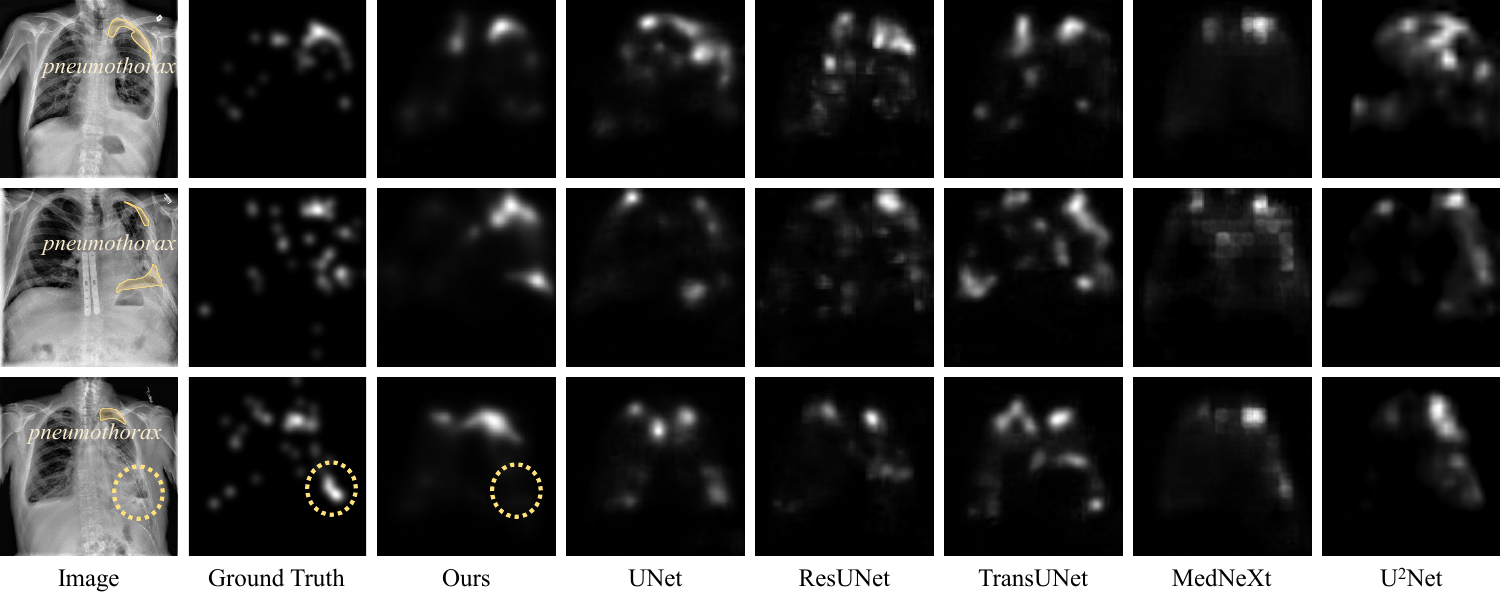}}
\vspace{-1.8mm}
\caption{Comparison of visual attention generated by various methods on the SIIM-ACR dataset. Compared to others, the visual attention generated by our method closely matches that of real radiologists. The yellow circle highlights an area missed by our method that is non-abnormal on the X-ray, underscoring that VCC-Net predicted visual attention is highly aligned with actual abnormalities.}
\label{fig:Result_SIIM_gaze}
\vspace{-4mm}
\end{figure*}

Table.\ref{tab:tab_SIIM} compares our method quantitatively with others on the SIIM-ACR dataset. 
VCC-Net significantly outperforms competing approaches in Acc, AUC, and F1. 
Specifically, compared to the best-performing EG-ViT, our model improves Acc by 2.80\% (88.40\% vs. 85.60\%).
Furthermore, we observe a substantial improvement in the F1, where our method outperforms GazeGNN (which utilizes VC during inference) by 2.48\% (88.08\% vs. 85.60\%).
Notably, unlike GazeGNN, our approach generates visual attention through the VAG, enhancing the model’s usability.
Table.\ref{tab:tab_EGD} presents a quantitative comparison of the EGD-CXR dataset. 
VCC-Net outperforms the first two categories of methods in  Acc, AUC, and F1. 
When compared to the best-performing GA-Net, our method improves Acc by 5.61\% (85.05\% vs. 79.44\%) and F1 by 5.67\% (84.87\% vs. 79.20\%).
These results confirm that VCC-Net effectively leverages VC to guide and supervise the network. 
When compared to GazeGNN, our method achieves slightly lower AUC (91.52\% vs. 92.30\%) but outperforms GazeGNN in Acc (85.05\% vs. 83.18\%). 
This discrepancy may be attributed to GazeGNN's use of real VC during inference, which enhances sample differentiation by incorporating radiologists' semantic information, whereas our method autonomously generates VC. Nevertheless, our AUC remains the highest among methods in the first two categories.

\begin{table}[!t] \small
    \centering
    \caption{ 
    Quantitative comparison between our method and other methods on the TB-Mouse dataset. Bold values represent the best results. \\
    }
    \vspace{-3mm}
    \setlength{\tabcolsep}{8pt}
    \begin{tabular}{p{80pt}|p{25pt}p{25pt}p{25pt}}
    \hline
    
    Method    & \textbf{Acc$\uparrow$}  & \textbf{AUC$\uparrow$}  & \textbf{F1$\uparrow$}  \\
    
    \hline

    \textcolor{blue}{$\spadesuit$} ResNet18 \cite{ResNet}
    & 83.73  & 92.23 & 83.75 \\
    \textcolor{blue}{$\spadesuit$} ResNet50 \cite{ResNet}
    & 88.95  & 95.87 & 88.96 \\
    \textcolor{blue}{$\spadesuit$} ResNet101 \cite{ResNet}
    & 87.86  & 96.06 & 87.80 \\
    \textcolor{blue}{$\spadesuit$} ViT \cite{ViT}
    & 88.23  & 95.64 & 88.20 \\
    \textcolor{blue}{$\spadesuit$} SwinT \cite{SwinT}
    & 88.18  & 94.95 & 88.05 \\
    \textcolor{blue}{$\spadesuit$} ViG \cite{ViG}
    & 88.40  & 95.17 & 88.33 \\

    \hline
   
    \textcolor{green}{$\blacklozenge$} M-SEN \cite{MSEN}
    & 89.86 & 96.55 & 89.85 \\
    \textcolor{green}{$\blacklozenge$} EML-Net \cite{EMLNet}
    & 90.64 & 97.28 & 90.25 \\
    \textcolor{green}{$\blacklozenge$} DeepGaze \cite{DeepGaze}
    & 90.45 & 96.68 & 90.42 \\
    \textcolor{green}{$\blacklozenge$} GA-Net18 \cite{GANet}
    & 87.18 & 94.65 & 87.15 \\
    \textcolor{green}{$\blacklozenge$} GA-Net50 \cite{GANet}
    & 91.05 & 96.76 & 90.98 \\
    \textcolor{green}{$\blacklozenge$} GA-Net101 \cite{GANet}
    & 90.59 & 96.99 & 90.48 \\
    \textcolor{green}{$\blacklozenge$} EG-ViT \cite{EGViT}
    & 89.55 & 96.61 & 89.56 \\
    \textcolor{green}{$\blacklozenge$}  \textbf{Ours}  
    & \textbf{92.41} & \textbf{97.84} & \textbf{92.41}   \\
    \hline
    \end{tabular}
    \label{tab:tab_TB}
     \vspace{-3.5mm}
\end{table}

\begin{table}[!t] \small
    \centering
    \caption{ 
    Quantitative results from ablation studies on VCC-Net components for the SIIM-ACR and TB-Mouse datasets. Bold values represent the best results. \\
    }
     \vspace{-3mm}
    \resizebox{8.0cm}{2.8cm}{
    \begin{tabular}{cccc | ccc}
    \hline

    \multicolumn{7}{c}{\textbf{SIIM-ACR}} \\
    \hline

    \textbf{$\mathcal{L}_{soft}$}  
    &\textbf{$\mathcal{L}_{hard}$}
    &\textbf{$\mathcal{L}_{aux}$}
    &\textbf{$\mathcal{L}_{align}$}
     & \textbf{Acc$\uparrow$}  & \textbf{AUC$\uparrow$}  & \textbf{F1$\uparrow$}   \\
    \hline
    
    \checkmark & & & 
    & 85.20 & 85.53 & 84.66 \\
    \checkmark &\checkmark & & 
    & 86.40 & 85.90 & 86.58 \\
    \checkmark & &\checkmark & 
    & 86.40 & 85.88 & 85.97  \\
    \checkmark &\checkmark &\checkmark & 
    & 86.80 & 86.91 & 86.32  \\
    \checkmark & & &\checkmark 
    & 87.20 & \textbf{87.59} & 87.37 \\
    \checkmark &\checkmark &\checkmark &\checkmark 
    & \textbf{88.40} & 86.12 & \textbf{88.08}\\
   
    \hline
    \multicolumn{7}{c}{\textbf{TB-Mouse}} \\
    \hline
 
    \textbf{$\mathcal{L}_{soft}$}  
    &\textbf{$\mathcal{L}_{hard}$}
    &\textbf{$\mathcal{L}_{aux}$}
    &\textbf{$\mathcal{L}_{align}$}
     & \textbf{Acc$\uparrow$}  & \textbf{AUC$\uparrow$}  & \textbf{F1$\uparrow$}   \\
    \hline
    
    \checkmark & & &   
    & 89.82 & 95.72 & 89.71 \\
    \checkmark &\checkmark & & 
    & 90.68 & 97.77 & 90.55 \\
    \checkmark & &\checkmark &  
    & 90.32 & 97.42 & 90.19 \\
    \checkmark &\checkmark &\checkmark & 
    & 91.05 & 97.17 & 90.93  \\
    \checkmark & & &\checkmark 
    & 91.50 & 97.02 & 91.41 \\
    \checkmark &\checkmark &\checkmark &\checkmark 
    & \textbf{92.41} & \textbf{97.84} & \textbf{92.41}\\
    \hline
    
    \end{tabular}}
    \label{tab:tab_L}
     \vspace{-4.5mm}
\end{table}

We employed Grad-CAM\cite{GradCAM} to visualize attention maps of different methods on the SIIM-ACR dataset, as shown in Fig.\ref{fig:Result_SIIM_grad}.
The first column shows the original image with the yellow mark indicating the pneumothorax region, and the red area highlighted in the other columns shows the areas the model focuses on.
Methods without VC tend to focus on irrelevant regions, and even methods with VC, such as GazeGNN and GA-Net, struggle to localize abnormal areas accurately.
Our model mitigates this issue, focusing more precisely on the pneumothorax region, thereby enhancing interpretability.
In normal chest X-rays, VCC-Net correctly identifies the costophrenic angle, a critical anatomical landmark.
Fig.\ref{fig:Result_SIIM_gaze} compares the visual attention with that from other methods. 
Our visual attention aligns well with the ground truth, although some regions (highlighted by the yellow circle) are missed. These areas do not show abnormalities in the image. 
Results suggest that VAG generates high-quality visual attention, further improving model transparency.

\begin{figure*}[ht]
\centerline{\includegraphics[width=17cm]{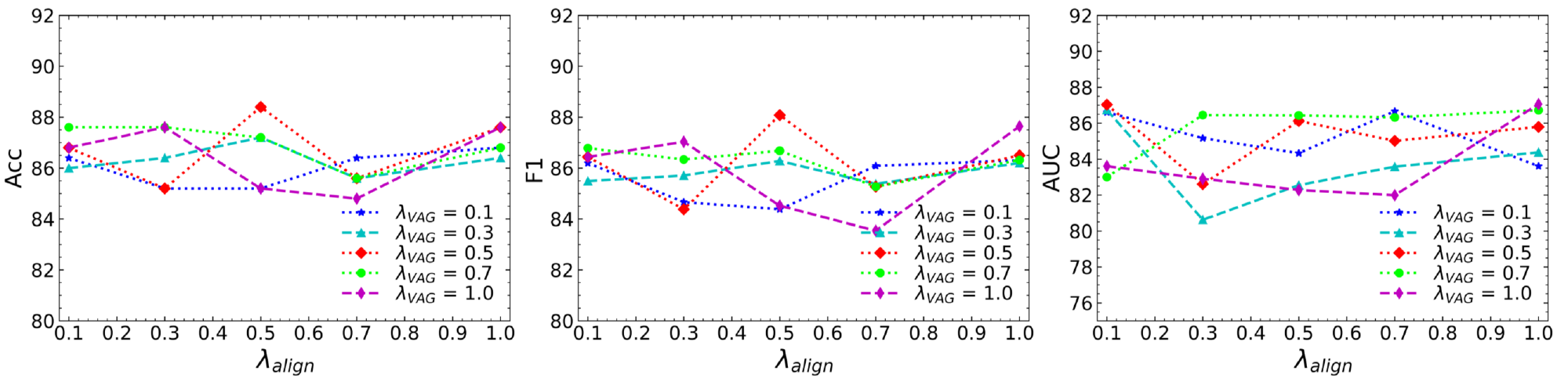}}
\vspace{-1.8mm}
\caption{Ablation study on the hyperparameters $\lambda_{align}$ and $\lambda_{VAG}$, which adjust feature alignment strength and the contribution of visual attention generation, respectively.
Acc, F1, and AUC for the SIIM-ACR data are shown across different parameter settings. Optimal performance is achieved when both parameters are set to $0.5$, with consistency observed across various settings, underscoring model stability.}
\label{fig:Result_lambda}
\vspace{-3mm}
\end{figure*}

\vspace{-2mm}
\subsection{Results on Mouse Dataset}
We compared our method against two categories of approaches on the TB-Mouse dataset: methods without VC \textcolor{blue}{$\spadesuit$} and with VC during training \textcolor{green}{$\blacklozenge$}. 
Since GazeGNN requires real VC, which is rarely available in practice, we did not compare our method to it.
Table.\ref{tab:tab_TB} shows the quantitative comparison on the TB-Mouse dataset, where our approach outperforms all others in Acc, AUC, and F1. 
Specifically, compared to the best-performing GA-Net50, our method improves Acc by 1.36\% (92.41\% vs. 91.05\%) and F1 score by 1.43\% (92.41\% vs. 90.98\%).
Fig.\ref{fig:Result_TB_grad} illustrates the attention maps on the TB-Mouse dataset generated by Grad-CAM for different methods.
The first column shows the original image with tuberculosis-related regions, such as exudation and nodules, marked with the bounding box.
Areas highlighted by the network are shown in red. 
Compared to other methods, ours more effectively identifies the abnormal regions.
Although ResNet detects some exudation areas, its localization is imprecise.
In contrast, our method accurately focuses on tuberculosis-related abnormalities, highlighting its enhanced interpretability.

\begin{table}[!t] \small
    \centering
    \caption{ 
    Quantitative ablation results on VAG structure for the SIIM-ACR and TB-Mouse datasets. Bold values represent the best results \\
    }
    \vspace{-3mm}
    \setlength{\tabcolsep}{8pt}
    \begin{tabular}{c | ccc}
    \hline
    
     \multicolumn{4}{c}{\textbf{SIIM-ACR}} \\
    \hline
    
    Method
     & \textbf{Acc$\uparrow$}  & \textbf{AUC$\uparrow$}  & \textbf{F1$\uparrow$}  \\
    \hline
    VCC
    & 83.20  & 84.89 & 82.81 \\
    VAG\,(CNN)\,+VCC
    & 85.60 & 82.94 & 85.14 \\
    VAG\,(GNN)\,+VCC
    & 87.20 & 85.07 & 86.14 \\
    VAG\,(GNN+CNN)\,+VCC
    & \textbf{88.40}  & \textbf{86.12} & \textbf{88.08} \\

    \hline
    \multicolumn{4}{c}{\textbf{TB-Mouse}} \\
    \hline
    
    Method
     & \textbf{Acc$\uparrow$}  & \textbf{AUC$\uparrow$}  & \textbf{F1$\uparrow$}  \\
    \hline
    
    VCC
    & 88.40  & 95.17 & 88.33 \\
    VAG\,(CNN)\,+VCC
    & 90.45 & 96.66 & 90.39 \\
    VAG\,(GNN)\,+VCC
    & 91.41 & 97.26 & 91.33 \\
    VAG\,(GNN+CNN)\,+VCC
    & \textbf{92.41}  & \textbf{97.84} & \textbf{92.41} \\
   
    \hline
    \end{tabular}
    \label{tab:tab_VAG}
    \vspace{-4mm}
\end{table}

\vspace{-2mm}
\subsection{Ablation Study}
\subsubsection{Effectiveness of Each Module of VCC-Net}
We conducted ablation studies on the SIIM-ACR and TB-Mouse datasets to evaluate the contribution of each component in VCC-Net, as shown in Table.\ref{tab:tab_L}.
When only the soft visual attention loss $\mathcal{L}_{soft}$ was used, the Acc was 85.20\% and 89.82\%, respectively.
Introducing $\mathcal{L}_{hard}$, which guides the model to focus on regions emphasized by radiologists, alongside the auxiliary classification loss $\mathcal{L}_{aux}$, led to an improvement in Acc.
Incorporating the alignment loss $\mathcal{L}_{align}$, which aligns feature differences with attention differences, resulted in a further Acc increase of 2.00\% (87.20\% vs. 85.20\%) and 1.68\% (91.50\% vs. 89.82\%) on the two datasets, respectively.
This demonstrates that $\mathcal{L}_{align}$ helps the model learn more effective representations by integrating cognition.
Finally, $\mathcal{L}_{align}$ was incorporated into both $\mathcal{L}_{hard}$ and $\mathcal{L}_{aux}$, leading to improvements in accuracy, which reached 88.40\% and 92.41\%, respectively.
A slight decrease in AUC was observed on the SIIM-ACR dataset. 
This may result from the hard visual attention, which tends to focus excessively on specific regions, potentially overlooking areas adjacent to anomalies.
While this enhances the model's ability to discriminate key features, it may reduce overall discrimination performance.

\subsubsection{Effectiveness of VAG's Architecture}
To investigate the effectiveness of the VAG architecture, we performed ablation experiments on the SIIM-ACR and TB-Mouse datasets, as shown in Table.\ref{tab:tab_VAG}.
Using only the original classifier, the Acc on the SIIM-ACR dataset was 83.20\%.
After generating visual attention with CNNs, Acc improved to 85.60\%, showing that VCC can effectively incorporate VC to enhance model performance.
When visual attention was generated using GNNs, Acc further increased to 87.20\%.
Combining GNN and CNN for attention map generation yielded the highest Acc of 88.40\%, outperforming all other strategies. 
Paired $t$-tests confirmed the statistical significance of these improvements, with p-values for Acc and F1 lower than 0.05.
Experiment results support the effectiveness of VAG in integrating both global and local information for more refined attention map generation.


\subsubsection{Ablation of Hyperparameters}
To evaluate the impact of the hyperparameters $\lambda_{align}$, $\lambda_{VAG}$, and $\alpha$ on model performance, an ablation study was conducted on the SIIM-ACR dataset.
$\lambda_{align}$ and $\lambda_{VAG}$ control the contributions of aligning radiologists' visual attention with feature differences and generating visual attention maps, respectively. Fig.\ref{fig:Result_lambda} presents performance metrics for various parameter settings $(0.1,0.3,0.5,0.7,1.0)$. 
The model exhibited optimal performance when both parameters were set to 0.5, and showed strong consistency and stability across different configurations.
We analyzed the effect of varying $\alpha$ values $(0,0.1,0.2,0.5,1.0,2.0,5.0,10.0)$ on model performance during graph construction. When $\alpha$ was set too low (e.g., $0$), the model depended heavily on learned features, yielding a relatively low accuracy of 83.2\%. As $\alpha$ increased, performance improved progressively, reaching its highest accuracy of 88.4\% at $\alpha = 2$. Beyond that point, larger $\alpha$ values diminished the contribution of feature information, causing a decline in accuracy to 85.6\% at $\alpha = 10$.

\begin{table}[t] \small
    \centering
    \caption{ 
    Comparison of performance using different types of visual attention. Bold indicates the best results. \\
    }
    \vspace{-3mm}
    \setlength{\tabcolsep}{8pt}
    \begin{tabular}{c | ccc}
    \hline
    \multicolumn{4}{c}{\textbf{SIIM-ACR}} \\
    \hline
    Method
     & \textbf{Acc$\uparrow$}  & \textbf{AUC$\uparrow$}  & \textbf{F1$\uparrow$}  \\
     
    \hline
    Random & 82.40  & 82.64 & 82.64 \\
    Radiologists' VC & 86.00 & 84.84 & 85.08 \\
    VAG & 88.40 & 86.12 & 88.08 \\
    VAG+Radiologists' VC & \textbf{88.80}  & \textbf{88.56} & \textbf{88.54} \\

    \hline
    \multicolumn{4}{c}{\textbf{TB-Mouse}} \\
    \hline
    Method
     & \textbf{Acc$\uparrow$}  & \textbf{AUC$\uparrow$}  & \textbf{F1$\uparrow$}  \\
    \hline
    Random & 90.85  & 95.58 & 90.97 \\
    Radiologists' VC & 91.32 & 97.77 & 91.33 \\
    VAG & 92.41 & \textbf{97.84} & 92.41 \\
    VAG+Radiologists' VC & \textbf{92.64}  & 97.79 & \textbf{92.64} \\
    \hline
    \end{tabular}
    \label{tab:tab_F}
    \vspace{-4mm}
\end{table}

\subsubsection{The ablation of different types of visual attention}
\begin{figure}[t]
\centerline{\includegraphics[width=7.8cm]{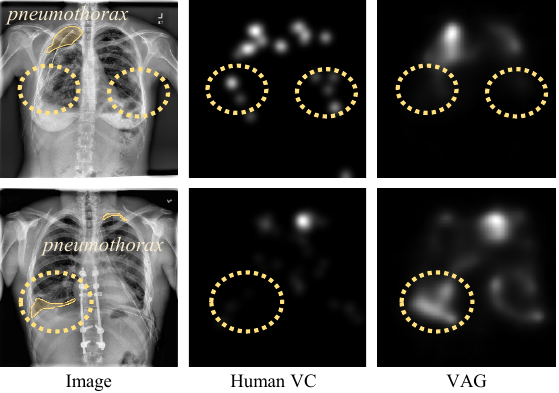}}
\vspace{-1mm}
\caption{Visualization reveals instances where radiologists, due to subjectivity or fatigue, misinterpret non-lesion structures as suspicious areas or overlook pneumothorax regions. The yellow-circled areas exemplify such occurrences.}
\label{fig:Result_attention}
\vspace{-4mm}
\end{figure}

Table.\ref{tab:tab_F} compares the performance of using different types of visual attention. Random refers to randomly sampled attention, whereas Radiologists' VC denotes visual attention derived from radiologists. The third type represents visual attention generated by the proposed VAG, and the fourth combines the generated and radiologists' VC through additive fusion. The attention produced by VAG achieved higher accuracy than VC on both datasets (88.40\% vs. 86.00\%, 92.41\% vs. 91.32\%). Such differences arise from the subjectivity and individual variability inherent in radiologists' vision, as radiologists occasionally fixate on non-lesion regions perceived as suspicious. Fig.\ref{fig:Result_attention} highlights this phenomenon, with yellow-circled areas illustrating examples of such attention. In contrast, VAG focuses more consistently on pathology-relevant regions, yielding modest yet measurable gains in diagnostic performance.
Integrating the model-generated visual attention with radiologists’ VC further improved accuracy relative to using predictions alone (88.80\% vs. 88.40\%, 92.64\% vs. 92.41\%). The results underscore a complementary relationship between radiologist and model. Radiologists’ attention provides clinically spatial constraints, while the model reduces subjective bias and optimizes focus distribution, enhancing diagnostic consistency. Overall, the findings reveal the promise of collaborative mechanisms within a human–AI collaborations and point toward the development of more reliable and efficient diagnostic models. 

\vspace{-2mm}
\subsection{Graph Structure and Distance Map Visualization}
\begin{figure}[t]
\centerline{\includegraphics[width=7.8cm]{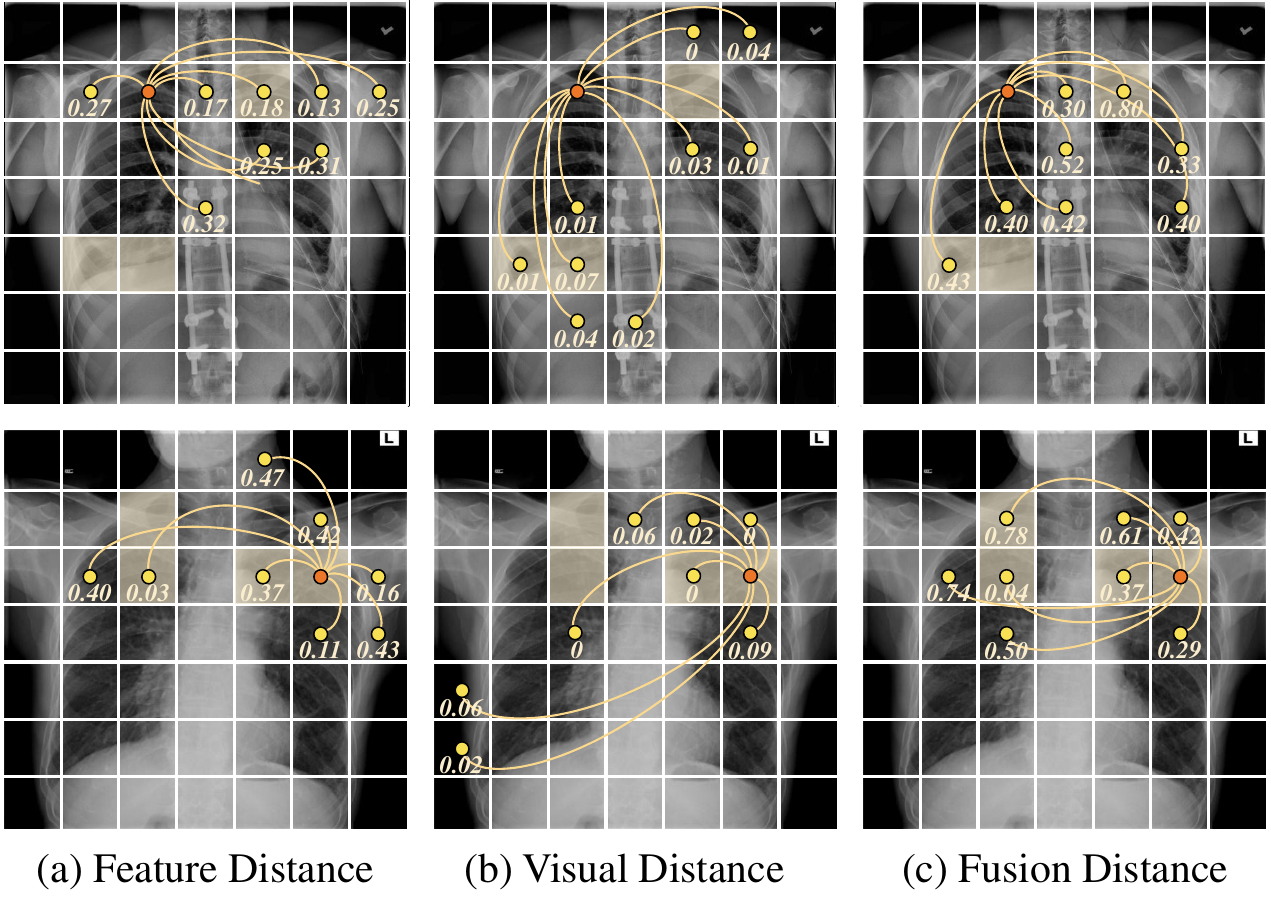}}
\vspace{-1mm}
\caption{Visualization of graph structures under different distance metrics. Examples from the SIIM-ACR and TB-Mouse datasets are shown. The red circle represents the central node with yellow neighboring nodes. Yellow areas highlight abnormal regions.}
\label{fig:Result_graph}
\end{figure}

Fig.\ref{fig:Result_graph} demonstrates the complementary effect of attention distance and feature distance during graph construction.
Two cases come from the SIIM-ACR and TB-Mouse datasets, respectively.
Each image is based on a distinct distance metric, with the central node indicated by the red circle and its neighbors indicated by the yellow circle.
The numbers in Fig.\ref{fig:Result_graph} indicate the actual distance between the central node and that node.
The yellow areas represent abnormal regions.
Initially, the central node, influenced by both feature and visual distances, often connects to irrelevant regions (\textit{e.g.}, background areas of chest X-rays). 
However, after distance fusion, these extraneous connections are effectively removed. 
As a result, the neighboring nodes of the central node focus more on the lung field, concentrating on potentially pathological areas.
This confirms the complementarity of the two distance metrics, with the fusion strategy facilitating the construction of a graph structure rich in foreground nodes while reducing interference from irrelevant regions, thereby enhancing the transparency and interpretability of the model.

\begin{figure}[t]
\centerline{\includegraphics[width=8.6cm]{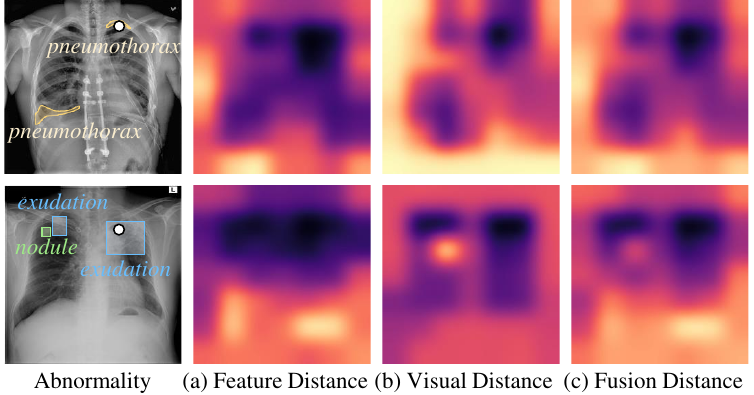}}
\vspace{-1mm}
\caption{Visualization of distance maps based on various distance metrics. Examples from the SIIM-ACR and TB-Mouse datasets are shown. Darker colors indicate proximity to the central white point.}
\label{fig:Result_distance}
\vspace{-4mm}
\end{figure}

Fig.\ref{fig:Result_distance} further illustrates the distance between the central white point and other areas in the image, with darker shades indicating shorter distances.
The two cases from the SIIM-ACR and TB-Mouse datasets display the chest X-ray along with abnormal pathological regions.
Subsequent columns show distance maps for different distance metrics.
In the first case, the central point (representing pneumothorax) is closer to other regions of the lung field in feature space, and the visual distance map successfully reduces the proximity to another pneumothorax area.
In the second case, the central point of the exudation region is closer to other exudation and nodules in feature space, but its boundary is blurred, erroneously bringing it closer to background areas.
The visual distance map increases the distance to unrelated regions, improving the construction of a more accurate graph structure.

\vspace{-2mm}
\section{Discussion and Conclusion}
This paper proposes the visual cognition-guided cooperative network (VCC-Net) to enhance the performance and interpretability of computer-aided diagnosis. VCC-Net comprises two components: visual attention generator (VAG) and visual cognition-guided classifier (VCC). 
VAG replicates the hierarchical visual search strategy of radiologists, generating attention maps to highlight critical regions. VCC constructs semantic associations between image regions and models inter-regional relationships using a graph structure, directing the model’s focus toward key lesion areas and aligning its decision-making with the radiologists' visual diagnostic process. Experiments on public gaze datasets (SIIM-ACR and EGD-CXR) and the self-developed TB-Mouse dataset, based on mouse trajectory, demonstrate that VCC-Net enhances diagnostic accuracy and interpretability significantly. Future work will explore the generalization of VCC-Net to additional imaging modalities and disease contexts to improve its adaptability. Another promising direction involves incorporating a human-in-the-loop mechanism to enable real-time radiologist feedback during inference, thereby refining decision-making and increasing clinical applicability.





\vspace{-1mm}
\bibliographystyle{IEEEtran}
\bibliography{IEEEabrv,reference}
\vspace{-1mm}

\end{document}